# An End-to-End, Segmentation-Free, Arabic Handwritten Recognition Model on KHATT

Sondos Aabed, *Birzeit University*, *1190652@student.birzeit.edu*,
Ahmad Khairaldin, *Birzeit University*, *1190499@student.birzeit.edu*,
*Advised by Dr. Ahmed Abusnaina, aabusnaina@birzeit.edu*
*Birzeit, Ramallah, Palestine*

**Abstract**.

An end-to-end, segmentation-free, deep learning model trained from scratch is proposed, leveraging DCNN for feature extraction, alongside Bidirectional Long-Short Term Memory (BLSTM) for sequence recognition and Connectionist Temporal Classification (CTC) loss function on the KHATT database. The training phase yields remarkable results 84% recognition rate on the test dataset at the character level and 71% on the word level, establishing an image-based sequence recognition framework that operates without segmentation only at the line level. The analysis and preprocessing of the KFUPM Handwritten Arabic TexT (KHATT) database are also presented. Finally, advanced image processing techniques, including filtering, transformation, and line segmentation are implemented. The importance of this work is highlighted by its wide-ranging applications. Including digitizing, documentation, archiving, and text translation in fields such as banking. Moreover, AHR serves as a pivotal tool for making images searchable, enhancing information retrieval capabilities, and enabling effortless editing. This functionality significantly reduces the time and effort required for tasks such as Arabic data organization and manipulation.

## 1. INTRODUCTION

Arabic Language, as one of the oldest and most spoken languages in the world, has a rich handwritten heritage. It has 28 letters, some of which have the same structure but differ in an added dot, two, or three, the position of those dots may also vary. Each of the 28 letters has multiple ways of writing depending on its position in a word. Linked letters connect to the characters previous to and ahead of it, in the same word. For instance, the letter ب (Ba'a), when appears at the beginning of a word بحر (sea) or in the middle of a word حبر (Ink), or at the end of a word like in the word حب (Love). Separated letters are connected to the letters previous to it but not to the ones in front of it, in the same word. An example would be ر (Ra), where in the word حرب (War) it's connected to the letter ح (Hha) and separated from the letter ب (Ba'a).

The Handwritten Arabic has a cursive nature. Since Arabic characters can be linked together and there are many variations for each character, some handwriting styles can make it difficult to distinguish each character from another. This can lead to characters overlapping vertically, which means two characters may exist in the same vertical line as seen in Figure 1.1 below. In extreme cases, characters in different words may overlap. Due to these factors, there are 84 variations for Arabic characters. That introduces the challenge of segmenting the units of Arabic text into words, sub-words called ligatures (also called Part of Arabic Words PAWs [1] and characters. For example, the word ساحرة (Witch) is split into { ة، حر، سا } each element of the set is a sub word like حر (HaRa) and each of those are constructed using the 28 letters. All these unique characteristics make the handwritten Arabic challenging to recognize.

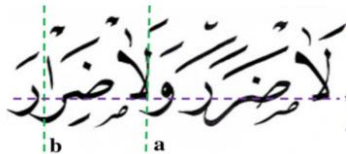

**Fig. 1:** Vertical overlapping (a) between two words (b) in one word [1]

Moreover, Arabic has faced challenges in digital recognition historically due to the characteristics discussed above. In addition to the hegemony of the English language in the industry, that resulted in marginalization towards other Languages. This linguistic bias has repercussions, especially in information retrieval from images. The significance of addressing such issues is a step towards contributing to language research in technology.

## 2. PROBLEM DEFINITION

In recent times, computer vision tasks have increasingly leveraged the power of deep learning and artificial intelligence. The problem of Handwritten Recognition is a computer vision classical problem. It has been approached and formulated differently over the years. In the specific context addressed here, the intelligent task involves replicating human-like abilities to read and recognize Arabic handwritten images. Which will be facilitated through a device Camera or Gallery just like other computer vision task. In this paper, the problem is formulated to be an *Image-based Sequence Recognition task*. It is described to be a sequence because the letters in reality occur sequentially as opposed to isolated characters. Figure 2 shows two examples of a sequence recognition problem. The task is to recognize a series of letters (labels) rather than a single label. That is, there is no need of having to detect each character and train on separate characters or perform any low-level segmentation which is considered as handcrafted feature extraction.

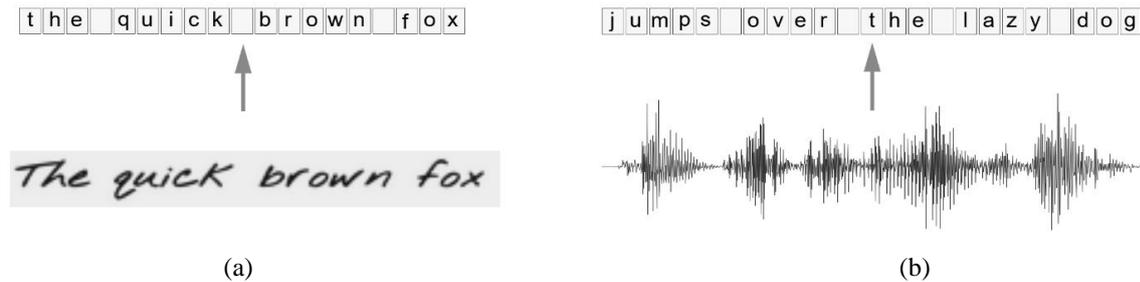

**Fig. 2:** Sequence Recognition Examples (a) Image-based recognition: The input can be (x, y) coordinates of a pen stroke or pixels in an image. (b) Voice-based recognition: The input can be a spectrogram or some other frequency-based feature extractor. [2]

## 3. PROPOSED METHODOLOGY

A segmentation-free framework for an image-based sequence recognition problem is proposed. The pipeline of the methodology starts with data analysis of the KHATT dataset, then moves to the data modeling with training experiments and finally gets to the Image Processing needed for doing inference.

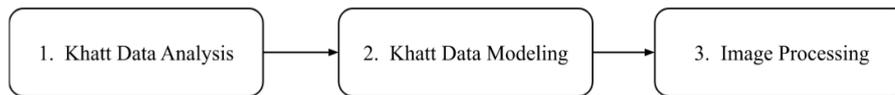

**Fig. 3:** Proposed Methodology Pipeline

### 3.1 KHATT Data Analysis

The dataset is loaded as assessments and observations are made, including distribution, outliers, or damages in both images and labels. For the preprocessing, the images are resized with a distortion-free resizing algorithm and the labels are padded with a maximum length of 70. Finally, data preparation uses the (70:15:15) splitting rule, and the data is batched with 16 batch sizes, moreover, caching and prefetching techniques are utilized for speeding data loading.

### 3.2 KHATT Data Modeling

The model architecture uses the *Deep Convolutional Neural Network (DCNN)* for Feature Extraction as the base model and two *Bidirectional Long-Short Term Memory (BLSTM)* for Sequence Classification. Dropout and batch normalization regularization techniques are utilized. The model is compiled on Adam Optimizer and a custom *Connectionist Temporal Classification (Loss)* based on the TensorFlow CTC implementation, in addition to the CTC decoder used in the prediction model. The Levenshtein Edit distance is used to calculate accuracy.

### 3.3 Image processing for inference

Input image processing is done in three phases: *Filtering*, *Transformation*, and *Line segmentation*. Filtering applied is Gaussian smoothing, total variation denoising, and then a gray-scaled image is obtained to get the binary image using Adaptive Local Thresholding. In terms of transformation, skew detection, and rotation using the Hough Algorithm. Distortion-free resizing that keeps the aspect ratio with padding as the second transformation technique. Finally, the high-level segmentation (Line Level) is approached using Horizontal Profile Projection and Path planning algorithms.

## 4. ETHICAL CONSIDERATIONS

Experiments within this project were conducted on data in a limited framework with no involvement of humans, animals, or other subjects. The only subject is the data which is the KHATT dataset, it was used under its license terms which is publicly available where the agreement conditions are agreed on KHATT Agreement 2014. The experiments conducted utilize Artificial Intelligence (AI) algorithms specifically deep learning which is under the UNESCO recommendations of ethical AI Recommendation 2021.

## 5. BACKGROUND

### 5.1 Relevant Theory

The Handwritten Recognition could be either offline or online recognition. Online means that the recognition is done in real-time, that's for instance when using smart pens or when writing with the fingers in the notes screen. Has software and hardware development. It is categorized into writer-dependent or writer-independent. The offline handwritten recognition, the data, in this case, is a static representation of the handwritten, for example, an image either uploaded or taken by camera or scanned by a scanner. It is categorized as character recognition (isolated) sometimes called Intelligent Character Recognition (ICR) or texts (Handwritten Recognition). Traditional Optical Character Recognition (OCR) and Handwritten Recognition are two technologies that significantly differ from each other, OCR is concerned with printed characters, template-based and it's ideal for fixed structured documents. There is also a difference between Handwritten and Handwriting recognition. While Handwritten Recognition is concerned with text recognition irrespective of the writer, the handwriting recognition task is concerned with who wrote the text, or giving it some context. For example, if it was written by a child or an adult, or if the text was written by a left-handed or right-handed individual.

A deep learning Pipeline for handwritten recognition is an end-to-end pipeline that searches for parameters to *minimize* the training loss, which is a differentiable function. These are too large parameters and training samples, so special optimizations were developed based on stochastic gradient descent (SGD) called mini-batch stochastic gradient descent. This introduces the step size parameter called Learning Rate that must be carefully adjusted with a schedule when training. Adam optimizer has recently been widely adopted because it utilizes both of the benefits of AdaGrad and RMSProp. The training loss function for sequence-based recognition tasks is the Connectionist Temporal Classification (CTC) Loss Function [2] which finds the maximum log-likelihood training of transcript, to integrate overall possible time-character alignments. For example: W="hi", T= 3 possible C such that K(C) = W: hhi, hii, __hi, h__i, hi_. The final objective is to maximize the probability of true labels (Multi-label classification). CTC is considered an alignment-free function, which collapses the repeats, also called many-to-one. Let's say the word cat is of length 3 shown in (a). It also makes a conditional independence assumption that implies that each output is considered independent of others given the input, which is a non-valid assumption in most sequence-to-sequence problems. In (b) If the prediction is 'A,' then the likelihood of the following sequence 'AA' should be significantly higher than that of 'riple A.' If 't' is predicted first, the opposite scenario should hold.

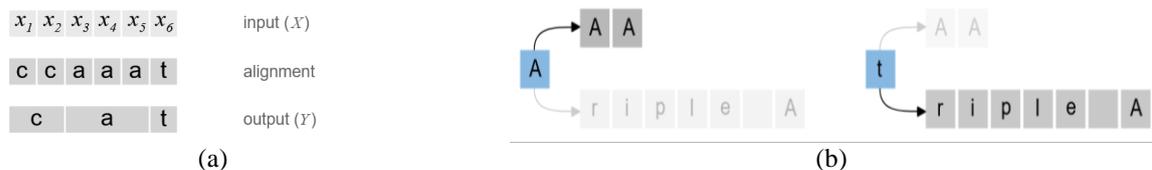

**Fig. 4**: Properties of CTC function (a) Alignment Free CTC (b) conditional independence assumption CTC

To measure the dissimilarity between predicted decoded sequences and true sequences *Edit distance Metris are used*. The goal is to minimize the edit distance during training to improve the accuracy of sequence prediction models. One notable algorithm is the Levenshtein distance algorithm it utilizes dynamic programming. The edit distance between حب (Love) and حرب (War) is one the insertion of one letter in the middle. However, if the data is not enough it would cause overfitting in deep learning. Some ways to help prevent overfitting to better generalization include **Dropout** which is a regularization technique where some percentage of the units are dropped to zero randomly for example 50%, to prevent memorizing. There is also, **Batch Normalization** which is to re-center the activations at a given unit so that they have unit variance and zero mean. [3]

Deep learning has different algorithms such as CNNs and RNNs. In Convolutional Neural Networks (CNNs) the idea is to convert each layer into feature maps and then combine them in different ways to get discriminative and semantically meaningful features. The number of learnable parameters in each layer is determined by the convolution window size (S), the number of input channels (C1), and the number of output channels (C2). The total learnable weights are S2 * C1 * C2. The computational cost per sample in a layer during forward and backward passes is WH * S2 * C1 * C2. The layer takes input from all Cin channels in the preceding layer, windows it to a small area, and generates values in one of the Cout channels in the next layer after the activation function. As for Long-Term Short-Term Memory (LSTM), Bidirectional means combining two independent LSTM cells which is a type of RNN, the first is processing the input from start to end, and the other one is from end to start and then concatenated. Standard RNNs have limitations in accessing a broad range of contextual information, that's because of the vanishing gradient problem, due to the intensity of inputs during the recurrent connections. That's how the Long-short Term Memory arises to overcome this obstacle. LSTM cells have a Cell State (long-term) part of the cell, it plays a role in what information to keep and what information to discard. The Hidden State (short-term) part of the cell, reflects the information at a particular time step and is responsible for predictions. LSTM uses gates to facilitate the flow like Forget, Remember, Input, and Output Gates. [3]

Image processing is any type of manipulation that happens pixel-wise, represented in the form of a function that returns the transformed value. The function takes one or more input images to produce an output image. These functions are categorized: *Filtering, Transformation, and Segmentation*. **Image filtering** is taking a kernel and applying it to the whole matrix of the image that does changes that affect the intensities of the pixels around that image with the image shape as it is. Its main aim is to denoising that image or create a noise in that image. Techniques like Gaussian smoothing and Total Variation Denoising are essential types of filtering, used in noise reduction, and structural edge preservation, by decreasing the total variation, unwanted detail is removed. There is also a Greyscale Color which focuses on the brightness variations of the red, green, and blue images. Finally, obtaining a binary image is a very simple representation of an image where the values are only (one or zero) using either Edge Detection or Thresholding. To obtain edges, two algorithms are famously used are canny and Sobel. As for Thresholding, there are Global and Adaptive or Local thresholding: While global applies the same threshold all over the image, adaptive (local) applies different thresholds on different parts of the image. Now ***Image Transformation*** takes an image or a set of images and reshapes it or makes changes on the dimensions of and matrix. Its main aim is to transform the visual representation of the image, such as rotation, and resizing to enhance features. Skew Detection and Rotation One of the algorithms used is Hough Transform-based skew detection, it assumes that text characters are aligned. Resizing is transforming the size with different algorithms such as distortion-free that keeps the aspect ratio. Finally, there is a tri-level ***segmentation***: line, word, and character segmentation. Modern approaches encourage segmentation-free. There are two methods for line segmentation, **Path Planning** method defines an optimal trajectory, analyzes pixel distribution, considers connectivity, adapts to image content, and iteratively explores paths. As for **Horizontal Histogram Profile Projection**, the number of foreground pixels or peaks within a row are counted to be the lines in that image.

### 5.2 Systematic Literature Review

The systematic review of Arabic handwritten datasets, summarized in Table 2.5, highlights key databases. The IFN/ENIT database by [13] is the most cited, featuring 26,459 images of Tunisian city names and handwriting samples, but lacks variations. CENPARMI by [14], focuses on Arabic Cheques with around 3000 real-world images and domain-specific vocabulary. SUST-ARG has limited diversity, including only male Arabic names. The KHATT database has extensive coverage, comprising 1000 forms from 46 sources, each with two paragraphs, and featuring 1000 writers with diverse characteristics.

**Table. 2:** Systematic Literature Review of Arabic Handwritten Dataset

| Database | Description | Limitations |
|---|---|---|
| **IFN/ENIT** [13] | Most cited database. Contains 26,459 Tunisian city names and handwriting samples | Lack of variations as it only includes names. |
| **CENPARMI** [14] | Gather about 3000 real-world Arabic Cheque images. Includes Arabic legal and courtesy vocabulary. | Lack of variations (domain-specific). |
| **SUST-ARG** | Contains only male Arabic names. | Lack of diversity male names |
| **KHATT** [15] | 2000 paragraphs from 46 different resources covering 11 subjects. Has the highest number of writers (1000) with diverse characteristics. Training has 9327 lines, 165,890 words and 589,924 characters | — |

The systematic literature review of Arabic handwritten recognition approaches, presented in Table 1, showcases various methodologies and their associated performances. Notably, the MD LSTM network combined with CTC, as explored by [4], achieved an impressive recognition rate of 91.4% on the IFN/ENIT dataset. Other techniques, such as the CNN-SVM architecture with dropout, reported error rates ranging from 2.09% to 7.05% on the HACDB and IFN/ENIT datasets, as demonstrated by [8]. These findings underscore the diversity of approaches and their varying degrees of success in addressing Arabic handwritten recognition challenges.

**Table. 1:** Systematic Literature Review of Arabic Handwritten Approaches

| Authors | Approach | Datasets | Performance |
|---|---|---|---|
| [4] | MD LSTM network and CTC | IFN/ENIT | 91.4% |
| [5] | MDLSTM on handcrafted features | IFN/ENIT | 88.8% |
| [6] | Segmentation of words into graphemes, BLSTM | IFN/ENIT | 22.3 % |
| [7] | LSTM, BLSTM, MDLSTM for Pashto handwritten text recognition | KPTI | 9.22% |
| [8] | CNN-SVM architecture with dropout | HACDB IFN/ENIT | 2.09%, 5.83%, 7.05% |
| [9] | RNN with four-layer bidirectional GRU, CTC, dropout | IFN/ENIT | 86.4% |
| [10] | CDBN framework with augmentation and dropout | HACDB, IFN/ENIT | 98.86%, 92.9% |
| [11] | MD LSTM-CTC preprocessing and augmentation | **KHATT** | 80.02% |
| [12] | CNN-RNN-based model for Persian handwritten | IFN/ENIT | 91.7% |

The techniques and outcomes of image processing methods are summarized in Table 3. These methods aim to enhance image quality for accurate recognition processes. Filtering techniques like Gaussian Filter and Median Filter, as explored by [16] have shown promise in improving image quality, despite challenges with unstable lighting conditions. Binarization techniques, by the KHATT database, offer solutions to low contrast, resulting in improved image quality for Arabic handwriting recognition. Furthermore, innovative approaches such as signature elimination improved accuracy.

**Table. 3:** Systematic Literature Review of Arabic Handwritten Image Processing

| Technique | Description | Outcome |
|---|---|---|
| **Filtering** [16] | Researchers used techniques like the Gaussian Filter, Median Filter, Low pass Filter, etc. | The results were bad due to unstable lighting. |
| **Binarization** | KHATT database addressed the binarization issue by transforming images to a more useful form, overcoming low contrast and data loss in binarization. | Improved image quality for Arabic handwriting recognition. |
| **Signature Elimination** [16] | Elimination of signatures from an image based on connected components and finding average threshold. | Improving recognition process and better noise elimination. |
| **Line Segmentation** [17] | The projection-based technique was successfully evaluated for Arabic printed text. | Effective for printed text. |
| [18] | Divide and conquer algorithms segmented the image into lines and processed them line by line. | Efficient approach for segmenting lines. |
| [16] | Hybrid used morphological operators and k-means. | Improved word segmentation. |

# 6. IMPLEMENTATION

## 6.1 KHATT Data Analysis

The dataset contains 132 columns, the first one is the image path, and the rest of the columns are the Latin representation of the Arabic letters in order on that image. These columns combined give the Annotation or the Labels of that image (Sentence). The shape of the data is (11399 Rows, 132 Columns). None of the data types were numeric except that column 132 was all empty and was dropped. Some image extensions in the path column are .tif while in the folder they are in the lossy .jpg extension. Hence the .tif in the data frame was replaced with the .jpg extension due to differences in data versions. Majority of images contain at least one letter. The maximum number of letters found in a single image was 130. There are 68 unique labels used for annotating the images, and each sentence concludes with a semicolon (;). Common Arabic letters have a high frequency, while others are comparatively low. This distribution is believed to accurately represent the usage of characters in Arabic words. For assessing data quality, various techniques were employed, concerning data types.

a. **Annotations (Labels):** Part of assessing data quality is verifying that the images have been annotated correctly. A sample was taken and the labels matched the images. Coma was considered in the annotation but it is out of context, the formations were not considered. On the other hand, there were 29 unannotated images found. After visualizing the unannotated it is noticed that some of them are scribbles one of them is either a signature or a doodle. One of the unannotated is just a dot, another is two lines. One of them also looks like a signature. But some of them are human-readable line images. So it is not known yet why they were left unannotated. The following figure shows some of them:

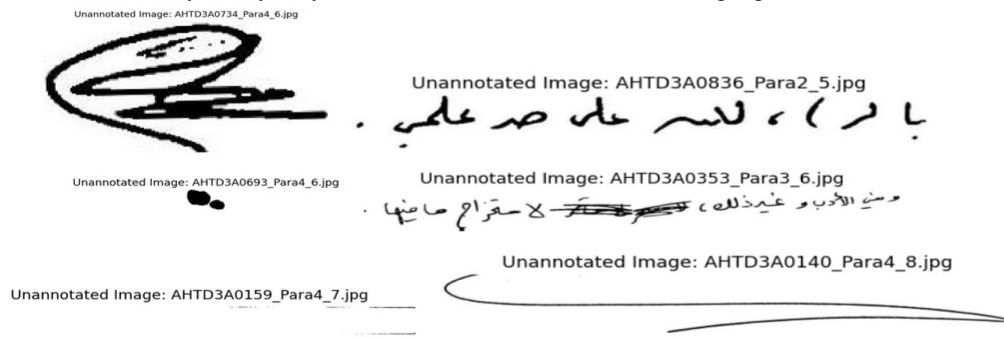

**Fig. 5:** Unannotated images sample

For the Unannotated, there are three options: either manually annotate the annotatable and delete the non-annotatable. Or delete all the 29 since the unannotated could confuse the model. Or even use them in testing the model inference. It is decided to save those images for testing the model inference and they are dropped from the data frame. On average most sentences have 67 letters. Most of the sentences have at least half of the maximum number of letters which is 130 here. Sentences that are longer than 70 are outliers. These NaN were padded with one value for consistency.

The label preparation starts by extracting the unique characters in the label columns, excluding null values, and creating a set named 'characters'. A mapping is also defined from Latin characters to Arabic characters to be used to convert specific Latin characters to their Arabic counterparts. Then String Lookup is used to create a mapping from characters to numeric values. Finally, a reverse mapping from numeric values to characters is also created. The data transformation starts by concatenating the label columns into a single column with spaces between characters (sequence). The maximum length of the labels is defined for the CTC layer. For the input length of the sequence, it is padded with a padding token. Let's say we have the sequence حرب (War) HaRaBa this is the corresponding label each character will be joined by a space ' ' to become the sequence Ha Ra Ba and then they are encoded using char to num and become[ 5 10 2 ] for example and let's say the padding token is 1 and the max length is 100 character (after dropping outliers), the sequence will be paddled until it reaches the length 70 [5 10 2 99 99 99 99 …etc.] The number of unique characters is 69 +1, [UNK] or unknown word so the padding value must always be larger than 69 + 1. Arabic is written from right to left (RTL) the labels must be read accordingly, that is why a reversing operation is done, and the sample becomes: [99 99 99 … 2 10 5]. The final step is to prepare data for training by splitting and loading it into batches. The chosen split rule is (70:15:15 rule) 70% of the data will be dedicated to training, while 15% each will be allocated for validation and testing purposes. Batching is implemented with a batch size of 16, grouping samples for efficient processing. Moreover, caching and prefetching techniques are employed to enhance data loading speed and reduce I/O bottlenecks.

b. **Images**: Regarding images, both sizes and aspect ratios were observed to follow a normal distribution. While there are 59 outliers in terms of image size, it was deemed appropriate to retain them, as all images will be resized to a consistent dimension when input into the AHR model. The overall sample indicated no instances of file damage; all files were successfully accessed. The following presents a general sample visualization of KHATT line images. Across six different lines, numbers, commas, and semicolons are observed, exhibiting variability in writing style. The letters are highly readable, suggesting a human-like quality. It is also observed that the images are binarized.

**Fig. 6:** KHATT data sample of lines

Regarding the images aspect ratios and sizes, those were handled with resizing all the images into a consistent size and shape (height, width, channel): (64, 800, 1). Each image instance is decoded to PNG format and resized to (64, 800) using a distortion free resizing custom function. For the normalization purpose, the image data is converted to a floating-point format and scaling pixel values to the range [0, 1]. The following figures shows the difference after visualizing a resized sample:

**Fig. 7:** Distortion-free resizing for line image (800, 64, 1)

**6.2 KHATT Data Modeling**

The adopted training method is to train the entire model from scratch; both the feature extractor and classifier are trained from scratch, without relying on any pre-trained models. The DCNN models are incapable of performing recognition on Sequence Image-based Recognition tasks. They mostly will fail to generalize unless there is a huge amount of data that covers most sentences in Arabic which is not possible at this moment. Attempts have been made on this approach to perform low-level segmentation for words, ligatures, and characters and then use DCNN on them (Character Recognition). This was the initially suggested algorithm, but implementations of low-level segmentation have not resulted in good accuracy in obtaining the low-level units, hence the decision to transition to a segmentation-free framework that is introduced. Now the problem is approached as a multi-label classification, the DCNN architecture is used for feature extraction and squeezed to be passed to Bidirectional Long-Short Term Memory (BLSTM) cells for recognition, on Connectionist Temporal Classification (CTC) Loss in addition to regularization techniques. To decode the predictions into final sequences of characters, a connectionist temporal classification (CTC) decoder layer is appended to the model. This decoder enhances the model's ability to recognize patterns in handwritten text by considering the temporal relationships between characters. With a total of 15,608,260 trainable parameters, the model exhibits robustness in capturing intricate patterns in the input data while facilitating efficient decoding.

Callbacks are also utilized, they are special utilities that are triggered when a certain event happens to prevent overfitting, logging training metrics, and reaching a better-trained model. Learning Rate Scheduler

(Custom implementation) schedules and changes the learning rate while training to find the best one. Edit Distance Callback (Custom implementation) to minimize the Levenshtein metric during training to improve the accuracy of sequence prediction models. Early Stopping (Keras implementation) to interrupt training if the passed validation loss remains unchanged for the last 10 consecutive epochs. Model Checkpoint (Keras implementation) saves the best model parameters. Finally, Log CSV (Keras implementation): Logs the training metrics to CSV files. The training is done on 50 epochs, and the batch size is 16. The results are pretty promising where the character correctly predicted is 84% and the Words correctly predicted is 71.29%. Even without Augmentation. Results are expected to reach State of the Art Model with Data Augmentation or Adaptive Data Augmentation. It is worth mentioning that these results were the highest on KHATT itself. The following are training and validation experiment results:

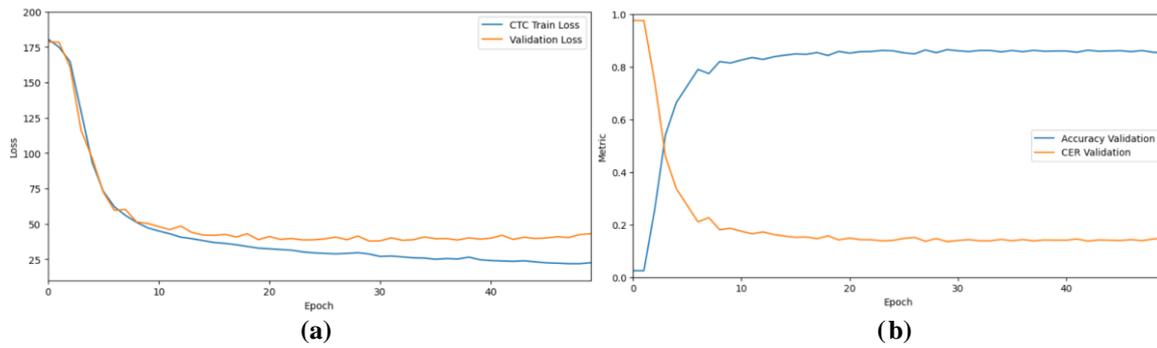

(a)          (b)

**Fig. 8:** Training results (a) CTC Loss on train as opposed to CTC loss on validation. (b) Accuracy and Character Error Rate per epochs.

### 6.3 Image Processing for inference

The reason why this phase is very important is that the audience of the AHR model could use different types of cameras, and in different environments where an image was taken the quality of the input image could be challenging. It's crucial to perform those techniques before doing an inference. There are three phases of image processing proposed: Filtering, Transformation, and Line Segmentation. Applying filters to reduce the noise, enhance quality, and get the gray color. The first filtering technique is Gaussian smoothing, it is used to smooth the edges and details. For denoising, a Total Variation Filter is applied to the smoothed image, and it makes the light more united. The Greyscale color image is gotten. To obtain a binary image local adaptive thresholding technique is applied. This binary image is used in the second phase of Transformation. To detect the most common angle in an image Hough Transformation method was used and then the image was rotated based on that angle. The other transformation technique is the image resizing using the same algorithm used for training which is distortion-free resizing. Finally, line segmentation is done using the horizontal profile projection with the threshold set to the mean.

The following samples are examples of doing inference on the data that is set for testing the model. It is noticed that the model has a great recognition rate with even law-represented data such as #. It is also noticed that the model has recognized some sentences with zero sequence error rates:

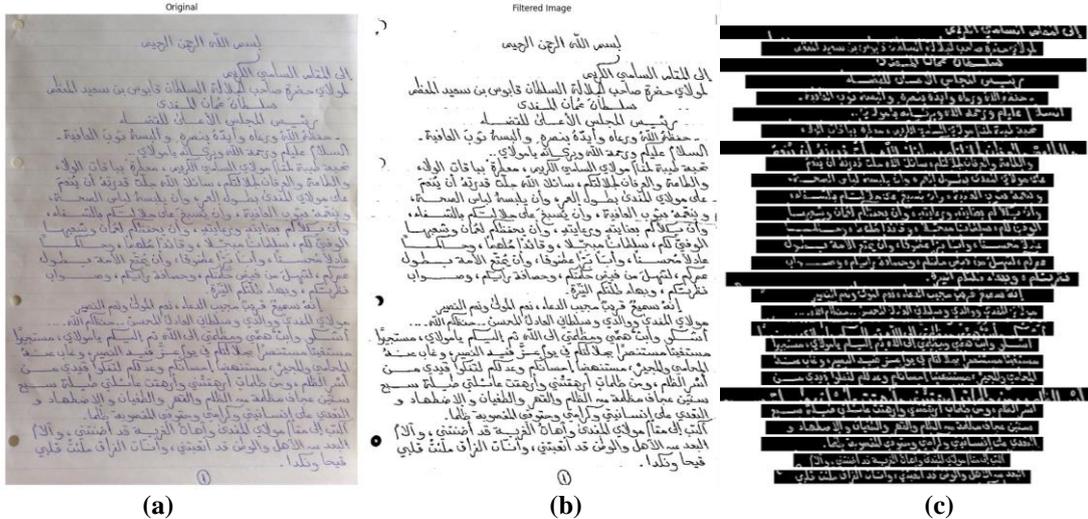

**Fig. 9:** Sequence Recognition Examples with AHR model

**(a)**                                  **(b)**                                  **(c)**

**Fig. 10:** Input Image Processing Phases (a) Original Image Scan, (b) Filtered Image (c) Transformation and Line Segmentation

**CONCLUSION**

In conclusion, this paper proposes a segmentation-free approach. Optimizations are proposed in distinct fields. For image processing, recognizing the challenges in accurate line segmentation due to the cursive nature of Arabic handwriting, a recommendation is made to explore training a model for paragraph recognition, eliminating the need for line segmentation. Data optimization focuses on addressing the observed data imbalance through adaptive data augmentation, anticipating substantial improvements in model performance. The extensive potential for model optimization involves investigating the Beam Search CTC Loss and experimenting with various optimizers, and activations. Additionally, considering an End-To-End approach with transformers and Encoders-Decoders could further contribute to refining the model's performance.

## ACKNOWLEDGEMENT

We offer our acknowledgment to the resilient and strong Palestinians, particularly those in Gaza, they all inspired hope and unity. This work is dedicated as a form of resilience in the face of the occupiers, for Palestinians wherever they are around the world. Special thanks to our beloved parents and family for their undivided support, which kept us fighting for victory.